\RequirePackage{fix-cm}
\documentclass[smallcondensed]{svjour3}     
\smartqed  
\usepackage{graphicx}
\usepackage{cite}
\usepackage{multirow}
\usepackage{amsmath}
\usepackage{amsfonts}
\usepackage{mathrsfs}
\usepackage[square,numbers]{natbib}
\usepackage[table,xcdraw]{xcolor}
\usepackage{latexsym}
\usepackage{textcomp}
\usepackage{amsmath}
\usepackage[table,xcdraw]{xcolor}
\DeclareMathOperator*{\argmin}{arg\,min}

\usepackage{hyperref}
\hypersetup{
    colorlinks=true,
    linkcolor=black,
    filecolor=black,      
    urlcolor=black,
    citecolor=black
}

\begin{document}

\title{11K Hands: Gender Recognition and Biometric Identification Using a Large Dataset of Hand Images
}

\titlerunning{11K Hands}        

\author{Mahmoud Afifi 
}


\institute{Mahmoud Afifi \at
              Electrical Engineering and Computer Science Dept., Lassonde School of Engineering, York University, Canada \\
              Information Technology Dept., Faculty of Computers and Information, Assiut University, Egypt
              \\
              \email{mafifi@eecs.yorku.ca - m.afifi@aun.edu.eg}           
}

\date{Received: date / Accepted: date}

\maketitle

\begin{abstract}
The human hand possesses distinctive features which can reveal gender information. In addition, the hand is considered one of the primary biometric traits used to identify a person. In this work, we propose a large dataset of human hand images (dorsal and palmar sides) with detailed ground-truth information for gender recognition and biometric identification. Using this dataset, a convolutional neural network (CNN) can be trained effectively for the gender recognition task. Based on this, we design a two-stream CNN to tackle the gender recognition problem. This trained model is then used as a feature extractor to feed a set of support vector machine classifiers for the biometric identification task. We show that the dorsal side of hand images, captured by a regular digital camera, convey effective distinctive features similar to, if not better, those available in the palmar hand images. To facilitate access to the proposed dataset and replication of our experiments, the dataset, trained CNN models, and Matlab source code are available at (\href{https://goo.gl/rQJndd}{https://goo.gl/rQJndd}).
\end{abstract}

\section{Introduction}

Biometric identification is the process of recognizing individuals using characteristics that can be behavioral or physiological \cite{nixon2008handbook}. Many studies show that hand dimensions possess distinctive features that can reveal gender information \cite{mcfaddenrelative, kanchananthropometry}. This kind of information is one of a set of attributes known as soft biometrics, which can be extracted from human biometric cues. Such soft biometrics can be integrated with primary biometrics traits to boost the accuracy of biometric identification \cite{surveysoftbiometrics}. An advantage of hand images is they are usually captured under a controlled position, unlike face images, which are usually unconstrained. Additionally, hands have less variability compared to, for example, faces, which are usually represented by deformable models due to facial expression changes \cite{surveysoftbiometrics, yu2016face}.
\\
Most state-of-the-art methods, which use hand images for either gender recognition or biometric identification, follow the traditional pipeline. Firstly, handcrafted features are extracted, followed by either training an off-the-shelf classifier or using a similarity measurement metric to compare such features with templates in the database \cite{surveysoftbiometrics}. However, recently, convolutional neural networks (CNNs) have been largely outperforming hand-engineered features-based methods. One drawback of the CNN, is its susceptibility to overfitting, if the CNN had been trained using a small dataset \cite{7328311}. On the other hand, the CNN can be useful even with a small dataset, as shown by Nanni \textit{et al.} \cite{nannihandcrafted}, on the effectiveness of using the trained CNN as a generic feature extractor - handcrafted features can be substituted by deep transfer learning features (we named it CNN-features), achieving highly accurate results for image classification. Interestingly, before being used as a feature extractor, these trained CNNs could be trained to solve another problem. 
\\
In this paper, we present a large dataset of hand images, which we believe to be an essential resource to ameliorate the development of gender recognition and biometric identification techniques.  As we have a large dataset of hand images, we are able to effectively train a CNN. To that end, we train a two-stream CNN for gender recognition using the proposed dataset. We then employ this trained two-stream CNN as a feature extractor for both gender recognition and biometric identification. The latter is handled using two different approaches. In the first approach, we construct a feature vector from the CNN-features to train a support vector machine (SVM) classifier. In the second approach, three SVM classifiers are fed by the CNN-features extracted from different layers of the trained CNN and one SVM classifier is trained using the local binary pattern (LBP) features \cite{1017623} in order to identify the person via his/her hand image. The reason for not training directly the CNN for the biometric identification task (i.e., end-to-end training) is that the hand images we have for each subject is not adequate to effectively train a CNN -- even though the average number of images per subject (58 images per subject) is higher than the average number of hand images per subject of the existing hand datasets in the literature.
\\\\
\noindent \textbf{Contribution:} \\The contribution of this paper can be summarized
as follows. 
\begin{itemize}
\vspace{-5mm}
\item A new dataset of more than 11,000 hand images is proposed. 
\item A two-stream CNN is presented for the gender classification task, outperforming the classification accuracies obtained by other CNN architectures, such as AlexNet and GoogleNet.
\item The trained CNN is used as a feature extractor to feed a set of SVM classifiers for biometric identification.
\end{itemize}

The rest of the paper is organized as follows. In Section \ref{related}, we briefly review the related work. An overview of the image classification problem is provided in Section \ref{bg}. The two-stream CNN is presented in Section \ref{method}. In Section \ref{dataset}, we introduce the proposed dataset and the evaluation criteria for gender recognition and biometric identification, followed by the experimental results obtained by the proposed method in Section \ref{results}. Lastly, the paper is concluded in Section \ref{conclusion}.

\section{Related work}
\label{related}
In the literature, local and global hand-engineered features were extracted from palmar hand images to feed a classifier or to measure the similarity between them and pre-extracted features for gender recognition and biometric identification tasks. In this section, we present a brief review of the previous gender recognition and biometric identification methods that rely on regular hand images (i.e., scanned or captured by a digital camera). Methods which depend on hand vein patterns obtained by infrared imaging are not included in the scope of this paper.
\subsection{Hand-based gender recognition methods}
 In 2008, Amayeh \textit{et al.} \cite{amayehgender} explored the possibility of extracting the gender information from images of the palmar side of the hands. They relied on the geometries of palm and fingers that were encoded using Fourier descriptors. To evaluate their method, they collected 400 palmar images of 40 subjects. After that, Xie \textit{et al.} \cite{xie2012study} used 1,920 images of the dorsal side of 160 hands to extract the skin texture patterns for gender classification and biometric identification. Wu and Yuan \cite{wugender} classified the hands' geometrical features using polynomial smooth SVM based on a separate 180 palmar images of 30 subjects. Unfortunately, these datasets are not available for further comparisons. 
\subsection{Hand-based biometric identification methods}
In the field of hand biometrics, geometrical and local features are extracted using generic or customized feature extractors, such as Radon transform, Gabor filter \cite{kumar2010}, scale invariant feature transform (SIFT) extractor \cite{charfi2014novel}, customized shape, geometrical, or figure feature extractors \cite{sharma2015identity,8032481}. The extracted features are used to drive the final decision, which is taken based on either a similarity metric or by an off-the-shelf classifier. 
For instance, Charfi \textit{et al.} \cite{charfi2014novel} used the cosine similarity as a similarity metric, while Sharma \textit{et al.} \cite{sharma2015identity} utilized the earth mover distance, $L1$, and Euclidean distance. The k-nearest neighbor (k-NN) classifier was used by Bera \textit{et al.} \cite{Bera2017}.
Charfi \textit{et al.} \cite{new2017} proposed a cascade architecture using a set of SVM classifiers. The SVM classifiers were trained using two different representations. The first being the traditional SIFT features extracted from the palmar hand and palmprint images, and the second being a sparse representation (SR) of these features.

\section{Background}
\label{bg}
The classification problem can be generally described by the following optimization problem: 
\begin{equation}
\label{Eq:classification}
 \underset{\pmb{\theta}}{\argmin}
 \frac{1}{n}\sum_{i=1}^{n}\mathop{\mathcal{L}}(l_i,\mathbb{F}(\mathbf{X}_i)),
\end{equation}
\noindent
where $\mathbb{F} \colon \mathbf{X}\rightarrow\hat{l}$ is a classifier that gives the estimated class $\hat{l}$ of features $\mathbf{X}_i=(\mathbf{x}_{i_1},\mathbf{x}_{i_2},...,\mathbf{x}_{i_N})$ for a given training data which contains $n$ records, and $\pmb{\theta}$ is the parameters' vector of $\mathbb{F}$. By minimizing the loss function $\mathop{\mathcal{L}}$ with respect to $\pmb{\theta}$, $\hat{l}$ approaches to the true label $l$ and the misclassification error is reduced. 
\\
In recent years, use of the CNN has led to state-of-the-art results in many image classification problems. The power of the deep learning is, indeed, on the high degree of non-linearity provided by a series of activation functions of the hidden layers. The CNN consists of a set of stacked layers; each one has learnable parameters whose values are to be learned during the training phase, aiming to reduce error, which in that case, is the classification error. In other words, the minimization of Equation \ref{Eq:classification} is performed by adjusting the learnable parameters (represented by $\pmb{\theta}$) through the entire network producing strong features, at the last fully connected (fc) layer, to distinguish between different classes. 
\\
In a recent study, Nanni \textit{et al.} \cite{nannihandcrafted} showed that applying a dimensionality reduction post-processing on the extracted deep features from pre-trained CNNs (i.e., CNN-features), followed by training SVM classifiers, yielded good image classification accuracies compared to the results obtained using handcrafted features. This dimensionality reduction process is carried out, by definition, to convert the CNN-feature vector from the original space into a new space of a smaller set of linearly uncorrelated directions via the map function $f: \mathbb{R}^D \rightarrow \mathbb{R}^C\text{ : }C\ll D$, where
\begin{equation}
\label{PCA}
f(\boldsymbol{x})=\boldsymbol{x}\boldsymbol{W}+\boldsymbol{b}.
\end{equation}
$\boldsymbol{W}=[\boldsymbol{w}_{1}^T, \boldsymbol{w}_{2}^T, ..., \boldsymbol{w}_{C}^T]$ is the matrix of weight vectors $\boldmath{w}_j\in \mathbb{R}^D$, $\boldsymbol{x}\in\mathbb{R}^D$ is the input vector, and $\boldsymbol{b} \in \mathbb{R}^C$ is the bias vector. The benefits of this method are: 1) only SVM classifiers need to be trained without the need to retrain the CNN. 2) It does not require a very large number of training data, like the case of training CNNs, since we train only an SVM classifier. We conclude the previous discussion by emphasizing the following points:
\begin{itemize}
  \item The CNN-features can be effectively used instead of, or in addition to, the handcrafted features as input to an off-the-shelf classifier.
  \item By reducing the dimensionality of the CNN-features (e.g., the (1 $\times$ 4,096) CNN-feature vector extracted from the last fc of AlexNet \cite{alex}) before training the classifier, the classification accuracy can be boosted.
\end{itemize}

\begin{figure}
\centering
 \includegraphics[width=0.9\linewidth]{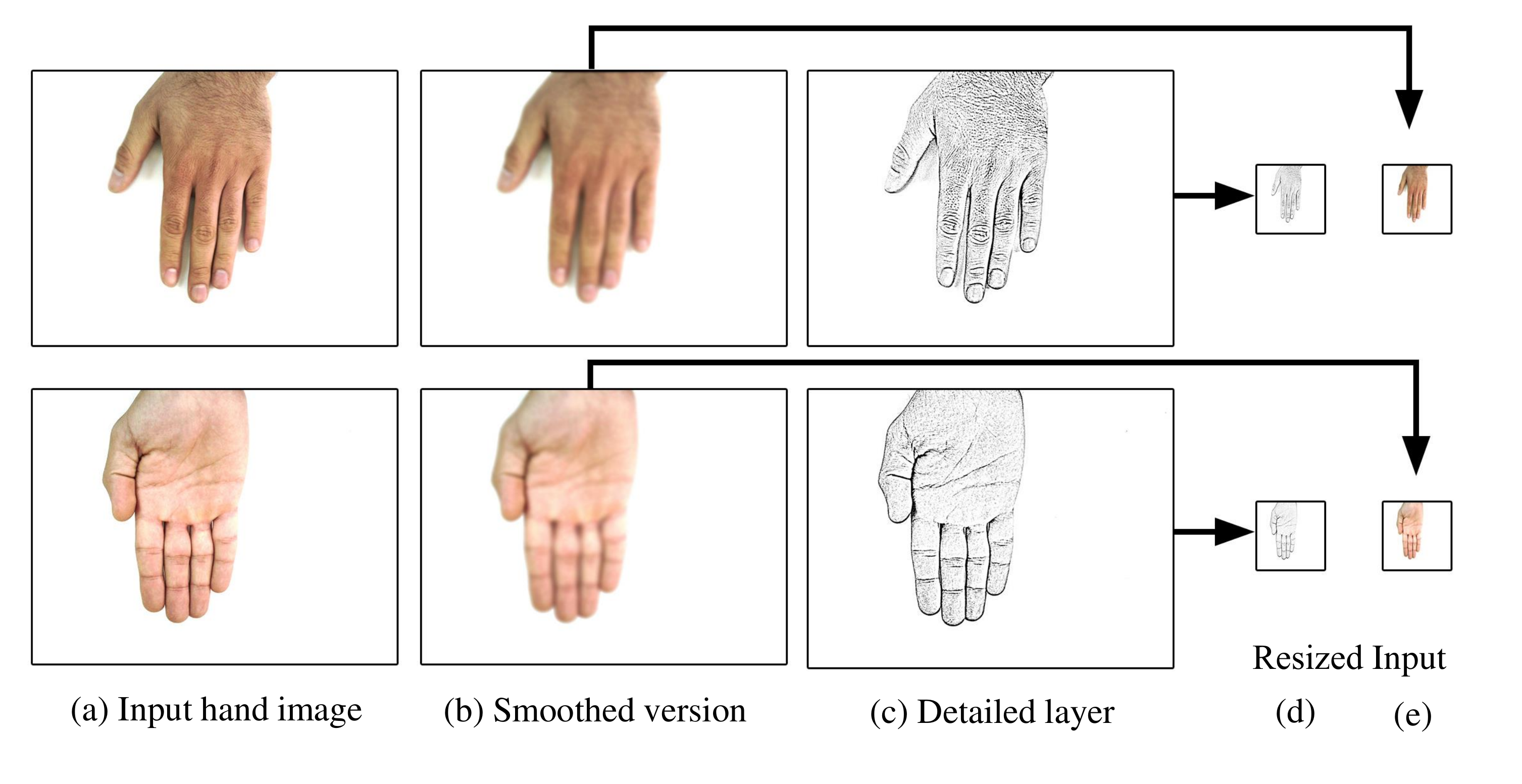}
   \caption{The pre-processing steps applied to the input images (a). The guided filter is applied to get the edge-preserving smoothed version of the original images as shown in (b). The high frequencies of the hand image are extracted and normalized in (c). Each image in (b) and (c) is resized to 224 $\times$ 224 pixels as shown in (d) and (e). Our CNN model receives images in (d) and (e) as input during the training and testing stages.}
    \label{fig:preprocessing}
\end{figure} 

\section{The proposed method}
\label{method} 
Gender recognition and biometric identification problems are largely considered a classification problem. The proposed method aims to tackle the aforementioned classification problems using the power of deep learning. We propose a two-stream CNN that is trained using two different inputs extracted from the hand images. A dimensionality reduction process is applied implicitly to the CNN-features generated by each stream, where this CNN will be used later as a feature extractor to train SVM classifiers in order to eliminate the problem caused by a lack of training data in biometric identification. To improve the accuracy of the biometric identification, the LBP is utilized to train another SVM classifier. This way of relying on training SVM classifiers creates the ability to expand the biometric system with new users without requiring a lot of computational time to retrain the CNN.

\subsection{Pre-processing}
\label{preproceesing_}
By applying edge-preserving smoothing for each image, we can extract the high frequency of the hand image (i.e., the detailed layer) through dividing the input image by the low frequency image (i.e., blurred or smoothed version of the input image), as proposed by Petschnigg \textit{et al.} \cite{flashnoflash}. One of the popular edge-preserving smoothing filters is the bilateral filter \cite{tomasibilateral}. It, however, suffers from limitations: 1) efficiency and 2) gradient reversal \cite{farbmanedge}. To solve that, we use the guided filter proposed by He
\textit{et al.} \cite{guidedF}, which is a linear time algorithm and obtains better results near edges compared to the bilateral filter.
\\
We apply the guided filter to get a smoothed version of the original hand image that will be used later to train the network. We find that the square window of the radius $r=10$ works well with our images, since each image is $1600\times 1200$ pixels as will be discussed in Section \ref{dataset}. The detailed layer is generated by a pixel-wise division as described in the following equation

\begin{equation}
I_{h}^{(x,y)}=\frac{I^{(x,y)}}{I_{l}^{(x,y)}+\varepsilon},
\end{equation}
where $I$ is the original hand image, $I_{l}$ is the smoothed image computed by the guided filter, $I_{h}$ is the detailed layer, $x$ and $y$ denote the spatial domain coordinates, and $\varepsilon$ is a small value for numerical stability. Eventually, the luma component (i.e., the ''brightness'' of the image) of the $\texttt{Y}^{'}\texttt{U}\texttt{V}$ version of the detailed layer is normalized, see Fig.  \ref{fig:preprocessing}. In other words, we discard the $\texttt{U}\texttt{V}$ channels (i.e., the chrominance components) of the hand image and employ only the brightness component $\texttt{Y}^{'}$ to represent the detailed layer.

\begin{figure*}
\centering
 \includegraphics[width=\linewidth]{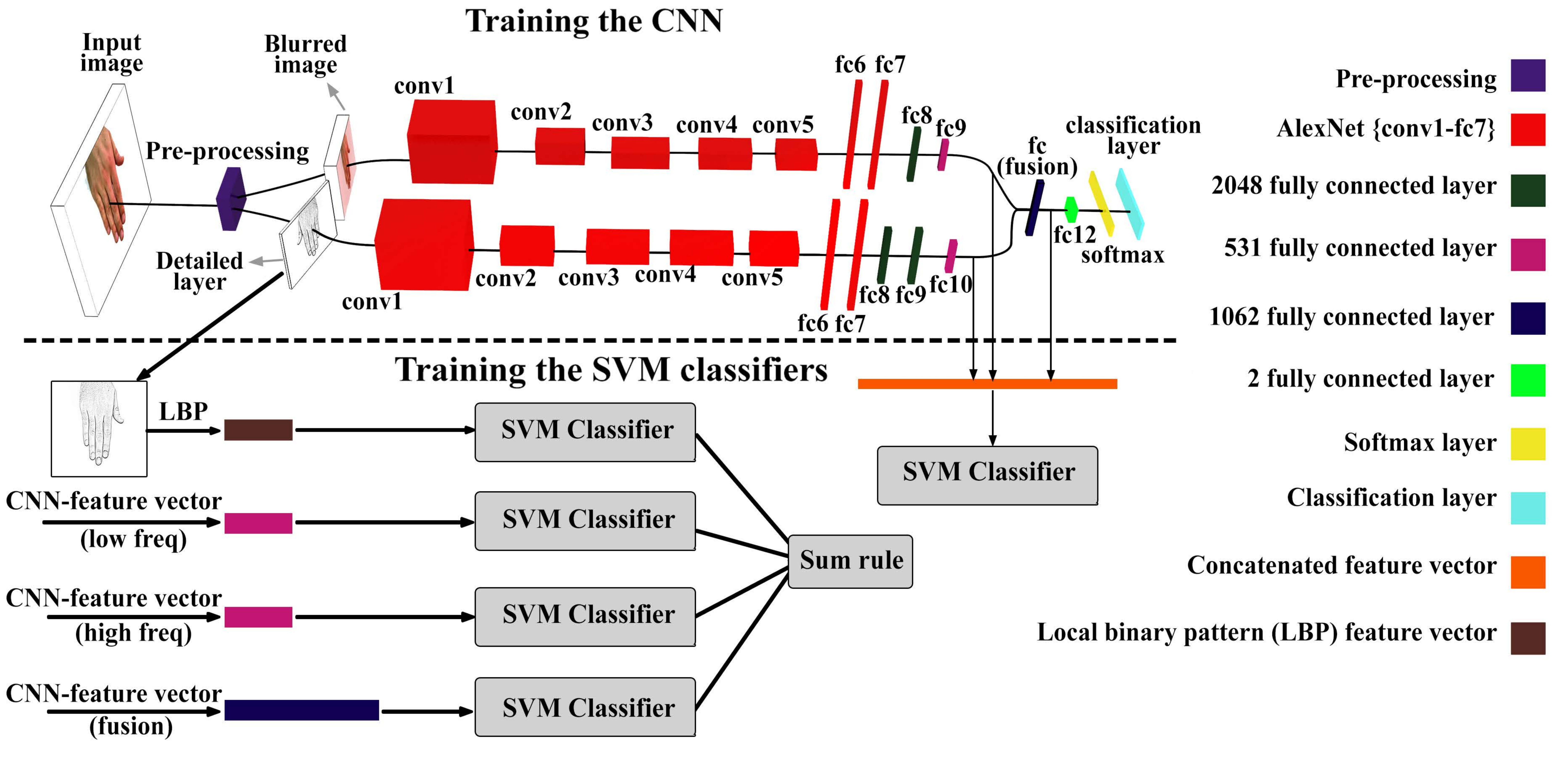}
   \caption{The proposed two-stream model. For the gender recognition, the two-stream CNN is end-to-end trained.  The blurred image constitutes the low frequency input and the detailed layer represents the high frequency input to the model. Each stream consists of 5 convolutional layers (conv) followed by a set of fully connected layers (fc) to extract the latent representation of the previous conv layer (i.e., conv5) and produce a deep feature vector at the last fc layer. After that, the trained CNN is used as a feature extractor; where an SVM classifier is trained using the concatenated feature vector extracted from the trained CNN. For the biometric identification problem, a set of SVM classifiers are trained using the extracted features from the last fc of each stream and the fusion fc layer. This fusion layer extracts a latent representation of the concatenated feature vectors of the last fc layers of the model's streams. Besides, we use the local binary pattern (LBP) feature. Eventually, the final score is computed by the sum rule of the output of each SVM classifier.}
    \label{fig:model}
\end{figure*} 

\subsection{The CNN architecture}
The proposed network consists of two streams which are combined in the last fusion layer, as shown in Fig. \ref{fig:model}. The first stream focuses on learning global features (e.g., hand shape and skin color, etc) by receiving the output of the guided filter. The second stream deals with the local features of the hand, provided by the detailed layer as described above. 
\\
At the beginning, we adopted the AlexNet architecture \cite{alex} for each stream and used the weights of the pre-trained AlexNet model as the initial values before the training process. However, there were some modifications performed to the AlexNet architecture. The first modification is applied to the second stream of our model; the weights of the first convolutional layer $\mathbf{w}_{conv1}$ of the second stream is initiated by computing the luma component of the first convolutional layer of the pre-trained AlexNet. This is due to the usage of the luma component of the $\texttt{Y}^{'}\texttt{U}\texttt{V}$ version of the detailed layer as input to the second stream of the proposed model. Instead of representing each of the 96 filters (also known as convolution kernel) as an ($11\times 11 \times 3$) tensor, we will represent each one as a ($121\times 3$) matrix $\mathbf{W}_{conv1}$ for convenience; such that, each column in $\mathbf{W}_{conv1}$ represents the filter coefficients of a color channel. The conversion is done by

\begin{equation}
\mathbf{w}_{conv1}=\mathbf{W}_{conv1}  \times \begin{bmatrix}
0.2989\\ 0.5870
\\ 0.1140
\end{bmatrix},
\end{equation}
where $\mathbf{W_{conv1}}$ is a ($121\times 3$) matrix containing all the weights of a 3D filter of the pre-trained AlexNet's first convolutional layer, and $\mathbf{w_{conv1}}$ is a ($121\times 1$) column vector representing the weighted sum of each corresponding weight of $\mathbf{W}_{conv1}$. 
\\
The remaining modifications are appending two fc layers and three fc layers to the first and second streams, respectively, to reduce the dimensionality of the output obtained from the fc7 layer of each stream. The output of each new fc layer is used as an input to an activation function and then a dropout layer is applied to prevent overfitting. For the last fc layer of each stream, there is no activation function applied nor dropout layer. For the first stream, the new fc layers (i.e., fc8, fc9) produce 2,048-dimensional and 531-dimensional outputs, respectively. For the second stream, the new layers, namely fc8, fc9, and fc10, produce 2,048-dimensional, 2,048-dimensional, and 531-dimensional outputs, respectively. For each new fc layer, except the last one at each stream, the output vector $\mathbf{y}$ is defined by 
\begin{equation}
\mathbf{y}= \sigma_{r} \circ f,
\end{equation} 
where $ \sigma_{r}(f)= max(f, 0) $ is the ReLU function and $f$ is the mapping function described in Equation \ref{PCA}. For the last fc layer at each stream, there is no activation function applied. It should be noted that, instead of applying a dimensionality reduction method (e.g., PCA) explicitly to fc7 of each stream, it is implicitly applied in the network using the learnable parameters $\mathbf{W}$ and $\mathbf{b}$ of the new fc layers.
\\
 Lastly, we append the fusion fc layer that receives the depth concatenation of the last fc layers of each stream. We apply average pooling to the output of the fusion layer with stride 2. The results are used as input to the last fc layer that has two output neurons. To classify, the softmax layer is used that predicts one of the two classes we have, namely, male and female labels. 

\subsection{LBP and CNN-features}
For the gender recognition, we test two approaches: 1) the two-stream CNN with a softmax layer, and 2) an SVM classifier that is trained using a concatenated CNN-feature vector. This vector contains the CNN-features from fc9 of the first stream, fc10 of the second stream and the CNN-features extracted from the fusion fc layer. 
We take this second approach and test it as one of the proposed approaches for biometric identification. Another proposed approach for biometric identification is based on a set of SVM classifiers, see Fig. \ref{fig:model}, in which three SVM classifiers are trained using CNN-features extracted from fc9 of the first stream, fc10 of the second stream, and the fusion layer, respectively. The fourth SVM classifier is trained using the LBP features extracted from the detailed layer. Eventually, the final score is computed by combining the output of all SVM classifiers by the sum rule. It worth noting that for the SVM classifiers used form the gender classification are a binary classifiers, while one-against-all SVM classifiers were used for the biometric identification.

\begin{figure}
\centering
 \includegraphics[width=\linewidth]{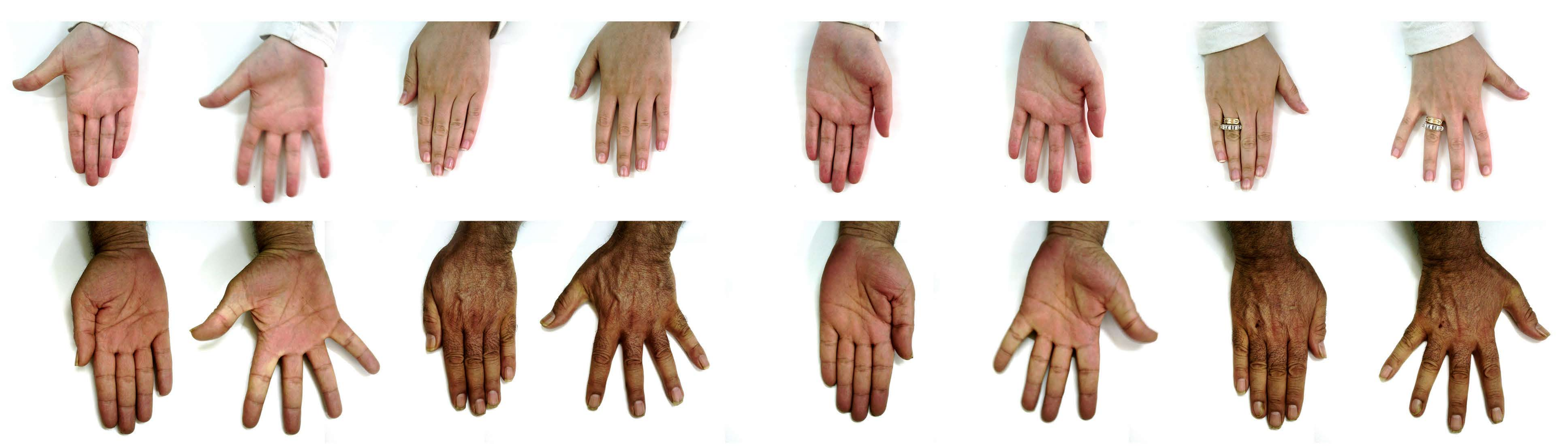}
   \caption{ Representative examples of the proposed dataset. Each row presents eight hand images of the same subject.}
    \label{fig:dataset}
\end{figure}

\section{The proposed dataset}
\label{dataset}
The proposed dataset, subsequently referred to as 11K Hands dataset, consists of 11,076 hand images (1600 $\times$ 1200 pixels) of 190 subjects, of varying ages between 18 - 75 years old. Each subject was asked to open and close his fingers of the right and left hands. Each hand was photographed from both dorsal and palmar sides with a uniform white background and placed approximately in the same distance from the camera (see Fig. \ref{fig:dataset} for representative examples of the proposed dataset). There is a record of metadata associated with each image which includes: 1) the subject ID, 2) gender, 3) age, 4) skin color, and 5) a set of information of the captured hand, i.e., right- or left-hand, hand side (dorsal or palmar), and logical indicators referring to whether the hand image contains accessories, nail polish, or irregularities. The proposed dataset has a large number of hand images with more detailed metadata, see Table \ref{Table1} for a comparison with other datasets. See Fig. \ref{fig:statistics} for the basic statistics of the dataset.

\begin{table*}[]
\centering
\caption{A comparison between the proposed 11K hand images dataset and other datasets.}
\label{Table1}
\scalebox{0.87}
{
\begin{tabular}{l|c|c|c|c|c|c|}
\cline{2-7}
 & \multicolumn{6}{c|}{{Dataset}} \\ \cline{2-7} 
 &  \begin{tabular}[c]{@{}c@{}}{11k Hands} \\ {(proposed)}\end{tabular} & \begin{tabular}[c]{@{}c@{}}{Sun \textit{et al.}}\\  {\cite{sunordinal}}\end{tabular} & \begin{tabular}[c]{@{}c@{}}{Yoruk} {\textit{et al.}}\\  {\cite{yorukshape}}\end{tabular} & \begin{tabular}[c]{@{}c@{}}{Hu \textit{et al.}}\\  {\cite{huhand}}\end{tabular} & \begin{tabular}[c]{@{}c@{}}{Kumar} \\ {\cite{kumarincorporating}}\end{tabular} & \begin{tabular}[c]{@{}c@{}}{Ferrer \textit{et al.}} \\ {\cite{ferrerlow}}\end{tabular} \\ \hline
\multicolumn{1}{|l|}{{Total images}} & \textbf{11,076} & 5,616 & 4,846 & 4,000 & 2,601 & 1,500 \\ \hline
\multicolumn{1}{|l|}{{Population}} & 190 & 312 & 642 (276) & 200 & 230 & 150 \\ \hline
\multicolumn{1}{|l|}{{\begin{tabular}[c]{@{}l@{}}AVG images \\ per subject\end{tabular}}} & \textbf{58} & 8 & 6 (3) & 20 & 11 & 10 \\ \hline
\multicolumn{1}{|l|}{{Hand side}} & \textbf{Palmar-dorsal} & Palm & Palmar & Palmar & Palmar & Palmar \\ \hline
\multicolumn{1}{|l|}{{Left-right}} & Both & Both & Both (left) & Left & Both & Right \\ \hline
\multicolumn{1}{|l|}{{Subject ID}} & Yes & Yes & Yes & Yes & Yes & Yes \\ \hline
\multicolumn{1}{|l|}{{Age}} & \textbf{Yes} & No & No & No & No & No \\ \hline
\multicolumn{1}{|l|}{{Skin color}} & \textbf{Yes} & No & No & No & No & No \\ \hline
\multicolumn{1}{|l|}{{Resolution}} & 1600$\times$1200 & 600$\times$480 & 382$\times$525 & 1382$\times$1036 & 1600$\times$1200 & NA \\ \hline
\multicolumn{1}{|l|}{{Gender}} & \textbf{Yes} & Yes & No & No & No & NA \\ \hline
\multicolumn{1}{|l|}{{\begin{tabular}[c]{@{}l@{}}Additional \\ details\end{tabular}}} & \textbf{Yes} & No & No & No & No & No \\ \hline
\end{tabular}
}
\end{table*}

\subsection{Hand images acquisition}
Each subject's hand was recorded using Point 2 View (P2V) USB Document Camera ($1600\times 1200$ pixels - 30 fps) with a constant brightness level of the camera under the same indoor lighting. After getting the recorded video, we measured the structural similarity between each consecutive pair of frames using the method proposed by Wang \textit{et al.} \cite{wangimage}.  We extracted only the frames that have a structural similarity below a predefined threshold to get rid of the redundancy in the recorded frames. We discarded any blurry, cropped, or oversaturated hand images after visual checks performed by 3 different persons. Hence, we obtained 58 images on average of each subject (30 dorsal and 28 palmar hand images on average).
\\
During the video recording sessions, the subjects answered a questionnaire to provide more information e.g., age, skin color, … etc. An annotator was present at all times to ensure correct image capture of hand sides as well as recording of subject ID. We then visually reviewed all of the metadata to avoid any errors that may have occurred in the initial annotation process.

\subsection{Evaluation Criteria}
\label{evaluation}
We exclude any hand image containing accessories from the training data to avoid any potential bias due to the visual information provided by these accessories in both problems-- gender recognition and biometric identification.

\subsubsection{Evaluation criterion for gender recognition}
\begin{figure}
\centering
 \includegraphics[width=0.65\linewidth]{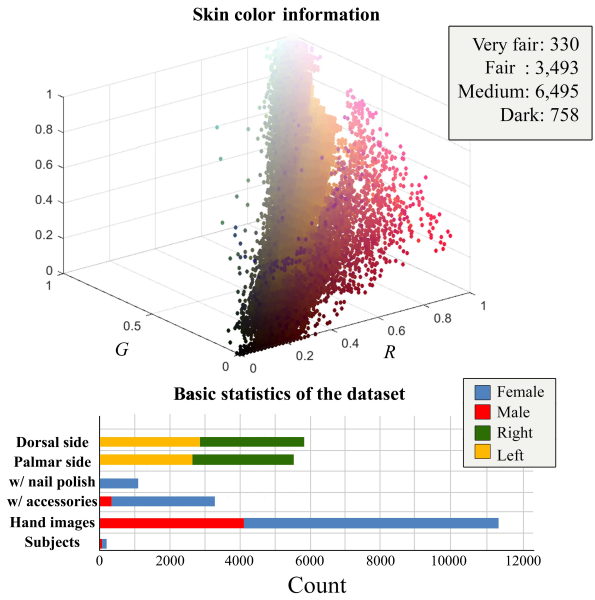}
   \caption{The summarized basic statistics of the proposed dataset. Top: the distribution of skin colors in the dataset. The number of images for each skin color category is written in the top right of the figure. The skin detection process was performed using the skin detection algorithm proposed by Conaire \textit{et al.} \cite{conairedetector}. Bottom: the number of  subjects, hand images (dorsal and palmar sides), hand images with accessories, and hand images with nail polish.}
    \label{fig:statistics}
\end{figure}

As we have a bias towards the number of female hand images, see Fig. \ref{fig:statistics}, we subsample 1,000 dorsal hand images of each gender for training and 500 dorsal hand images of each gender for testing. The images are picked randomly such that the training and testing sets are disjoint sets of subjects, meaning if the subject's hand images appear in the training data, the subject is excluded from the testing data and vice-versa. The same is done for palmar side hand images. For each side, we repeat the experiment 10 times to avoid overfitting problems and consider the average of accuracy as the evaluation metric. 
\subsubsection{Evaluation criterion for biometric identification }
For biometric identification, we work with different training and testing sets. We use 10 hand images for training and 4 hand images for testing of each hand side (palmar or dorsal) of 80, 100, and 120 subjects. We repeat the experiment 10 times, with the subjects and images picked randomly each time. We adopt the average identification accuracy (also known as correct identification rate) as the evaluation metric. Both training and testing sets for gender recognition and biometric identification are available on-line for further comparisons.

\section{Experimental results}
\label{results}
The experiments have been done on an Intel\textsuperscript{\textregistered} Xeon\textsuperscript{\textregistered} E5-1607 @ 3.10GHz machine with 32 GB RAM and NVIDIA\textsuperscript{\textregistered} GeForce\textsuperscript{\textregistered} GTX 1080 graphics card. 
\\
For gender recognition, we tested the proposed method against a set of image classification techniques which represent two main categories of image classification methods: 1) handcrafted feature-based methods, 2) CNN-based methods. For the handcrafted-based methods, we have tested the bag of visual words (BoW) framework \cite{csurkavisual} and the Fisher vector (FV) framework \cite{FV}. Both frameworks were introduced based on the SIFT features which are scale and orientation invariant. However, the SIFT features discard the color information that is considered by the CNN-based methods. Accordingly, we tested two color-based descriptors for the BoW and FV frameworks besides the SIFT descriptor. For the CNN-based methods, we have tested four different architectures beside our two-stream one. The experiments were done on the 11K Hands dataset using the aforementioned evaluation criteria discussed in Section \ref{evaluation}. 
\\
For biometric identification, the experiments have been conducted using our method on the proposed dataset and the IIT-Delhi-Palmprint (IITD) dataset \cite{kumarincorporating} to test the generalization of the proposed two-stream CNN as a feature extractor. Where, the hand images in the proposed dataset and the IITD dataset have different background, lighting conditions, and distances from the camera. The evaluation criteria described in Section \ref{evaluation} has been used for our dataset. For IITD dataset, we have used four images for training and one image for testing using 100, 137, and 230 subjects, as carried out in previous studies. We have reported the identification accuracy obtained using the features extracted from the CNN and the LBP features.
\\
In terms of performance, Table \ref{Table5} shows the time required for each step provided in this work. It should be noted that the training time of the SVM classifiers are quite small compared to the time spent training the CNN. That gives the ability to retrain the classifier to insert new subjects, in case of the biometric identification, without requiring high computational time. In the testing stage, the proposed method takes under one second to check a hand image allowing a real-time processing.

\begin{table}[t]
\centering
\caption{Time analysis in seconds of each process of the proposed method. 
The time shown is the average time required per image, except the CNN and SVM classifiers; where, the time shown represents the average training time required to train using all training set. All processes have been performed on CPU except the CNN-based operations carried out on GPU. 
}
\label{Table5}

\begin{tabular}{|l|c|c|}
\hline
Process & Training & Testing  \\ \hline

Pre-processing & - & 0.58863 \\ \hline

LBP feature extraction & - & 0.00857   \\ \hline
CNN feature extraction & - & 0.22523   \\ \hline
Two-stream CNN & 3168.016 & 0.01075 \\ \hline
SVM  (gender classification) & 1.0842 & 0.000007 \\\hline
SVM (biometric identification) & 114.5 & 0.00281 \\ \hline
\end{tabular}

\end{table}

\subsection{Training}

\begin{figure}
\centering
 \includegraphics[width=\linewidth]{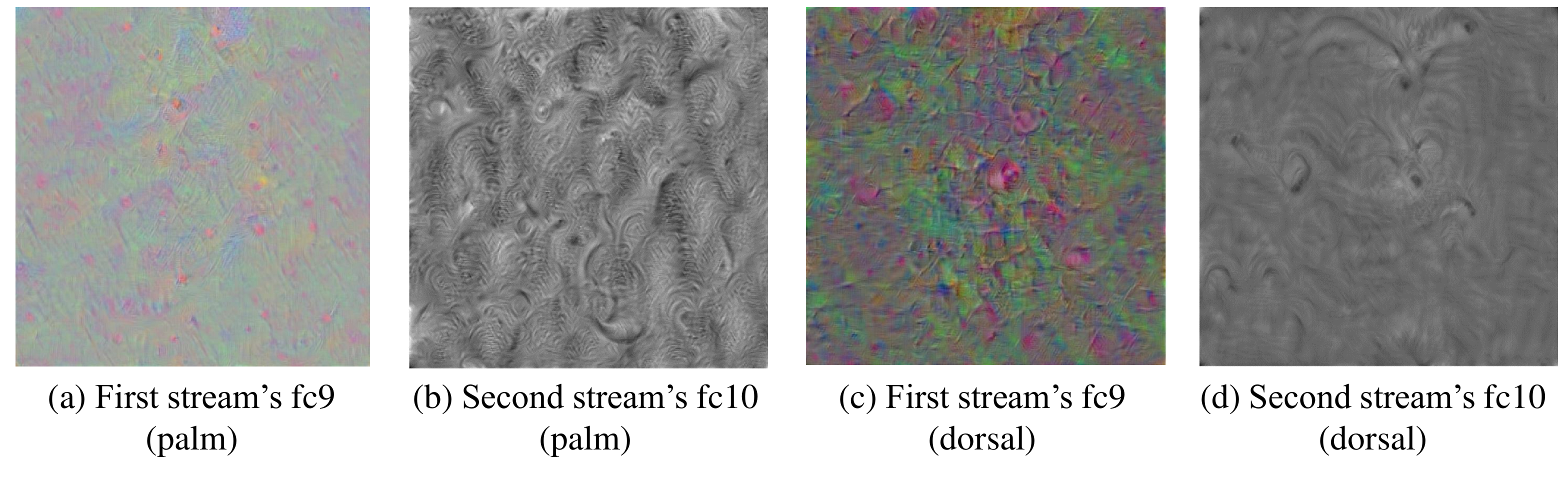}
   \caption{Feature visualization of the last fully connected layer of each stream, namely fc9 and fc10, using our trained network. (a) and (b) show feature visualization of fc9 and fc10 of our trained network on the palm side of the hand images. (c) and (d) show the visualization of the same layers of our trained network on the dorsal hand images.}
    \label{fig:deepdream}
\end{figure}

For CNN training, we use the stochastic gradient descent with momentum \cite{murphymachine} to update the parameters. We start by training each stream separately as an independent CNN by appending a fc layer with 2 output neurons followed by a softmax layer to each stream until convergence (typically 15-20 epochs  are sufficient). At this stage, we use the weights of the pre-trained AlexNet as the initial values of the learnable parameters. Then, the two-stream CNN is jointly trained until convergence (7-9 epochs are adequate) using the final weights computed in the first stage of training. In both stages, we use learning rate $\lambda=10^{-4}$ for the pre-trained layers and $\lambda=0.002$ for the new fc layers. Fig. \ref{fig:deepdream} shows feature visualization of the network after training; the shown images are synthesized after maximizing the activation of the last fc of each stream using deep dream \cite{mordvintsev2017deep}. This visualization shows the image features learned by our network for each side of hand images. As shown, each stream focuses on different patterns on the input hand image.
\\
For SVM training, we found that the linear SVM kernel is sufficient for gender recognition. However, for the biometric identification, a 2-d polynomial kernel is used in each SVM classifier. The optimization process is performed using the iterative single data algorithm \cite{kecman2005iterative}.

\subsection{Results of gender recognition}

Table \ref{genderEvaluation} shows a comparison of the average accuracy obtained by a set of generic image classification techniques and the proposed method. For BoW and FV frameworks, we have tested three different descriptors: 1) SIFT, 2) Color-invariant SIFT (CSIFT) \cite{dis_CSIFT}, and 3) \textit{rg}SIFT \cite{vanevaluating}. The number of visual words (i.e., clusters) was 200. For the CNN methods, namely, AlexNet, VGGNet (16 and 19) \cite{vgg}, and GoogleNet \cite{googleNet}, we have used learning rate $\lambda=0.002$ for new layers and $\lambda=10^{-4}$ for fine tuned layers. The training has been performed until convergence or reaching a maximum of 30 epochs using stochastic gradient descent with momentum. We had chosen 30 epochs as the maximum number of epochs to get a fair comparison between our model and others. Moreover, we found that 30 epochs are sufficient for all of the aforementioned architectures to converge. Table \ref{genderEvaluation} shows that all classifiers achieve higher gender recognition rates using the dorsal hand images compared to the obtained accuracies using the palm side of the hand images.

\begin{table}[]
\centering
\caption{Gender recognition results using our dataset. Each number represents the average accuracy.}
\label{genderEvaluation}
\scalebox{1.2}
{
\begin{tabular}{|l|c|c|}
\hline
  \multirow{2}{*}{Method} &  \multicolumn{2}{c|}{Hand side} \\ \cline{2-3}
   & Palm &   Dorsal \\ \hline
Bag of visual words/SIFT  &   0.680   & 0.727 \\ \hline
Bag of visual words/CSIFT &   0.670   & 0.819 \\\hline
Bag of visual words/\textit{rg}SIFT    & 0.703   & 0.814 \\ \hline
Fisher Vector/SIFT &   0.836   & 0.863 \\ \hline
Fisher Vector/CSIFT &   0.790   & 0.876 \\ \hline
Fisher Vector/\textit{rg}SIFT   & 0.794   & 0.877 \\\hline
CNN/AlexNet &   0.839   & 0.891 \\ \hline
CNN/VGG-16 &   0.862   & 0.907 \\ \hline
CNN/VGG-19 &   0.887    & 0.907 \\ \hline
CNN/GoogleNet &   0.879   & 0.889 \\ \hline
Ours (CNN)&    0.874   & 0.910 \\ \hline
Ours (SVM) &  \textbf{0.942}   & \textbf{0.973} \\ \hline
\end{tabular}
}
\end{table}

\begin{figure}
\centering
 \includegraphics[width=\linewidth]{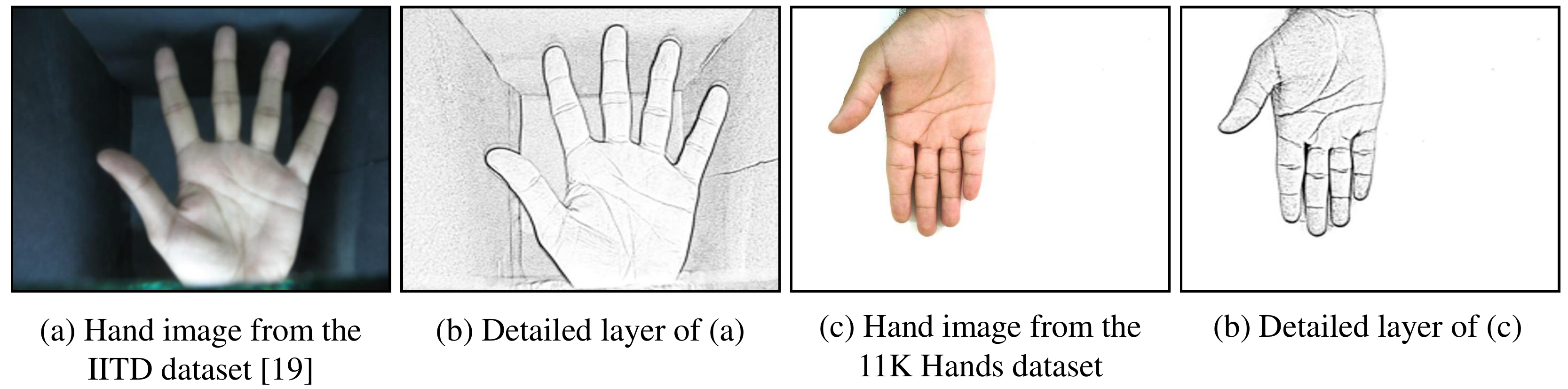}
   \caption{Two hand images from the IITD dataset \cite{kumarincorporating} and our 11K Hands dataset. (a) Input image from the IITD dataset. (b) The detailed layer generated of (a) as described Sec. \ref{preproceesing_}. (c) and (d) are a hand image from the 11K Hands dataset and the detailed layer of it, respectively. }
    \label{fig:comparison_IITD_ours}
\end{figure}

\subsection{Results of biometric identification}
The ensemble classifier, based on the four SVM classifiers, outperforms the single SVM classifier using the combined CNN-feature vector. Table \ref{Table3} shows the results obtained using different number of subjects. It is worth noting that the results obtained using the dorsal side of hand images outperforms the results of using the palmar hand images, similarly to the case of gender recognition.
\\
In order to validate the efficiency of the CNN-features extracted from the proposed CNN, we have tested both the single SVM and the ensemble classifier on the IITD dataset. Table \ref{Table4} shows the results obtained using the IITD dataset using the proposed technique against some of recent work used the same dataset. As shown in Table \ref{Table4}, we have achieved a good accuracy that approaches the results obtained by customized handcrafted feature extractors. Our results were obtained despite not retraining the proposed CNN on the IITD dataset and, most importantly, the environments of the captured hand images of each dataset are different in terms of hand position, background, and lighting, as shown in Fig. \ref{fig:comparison_IITD_ours}.

\begin{figure}
\centering
 \includegraphics[width=\linewidth]{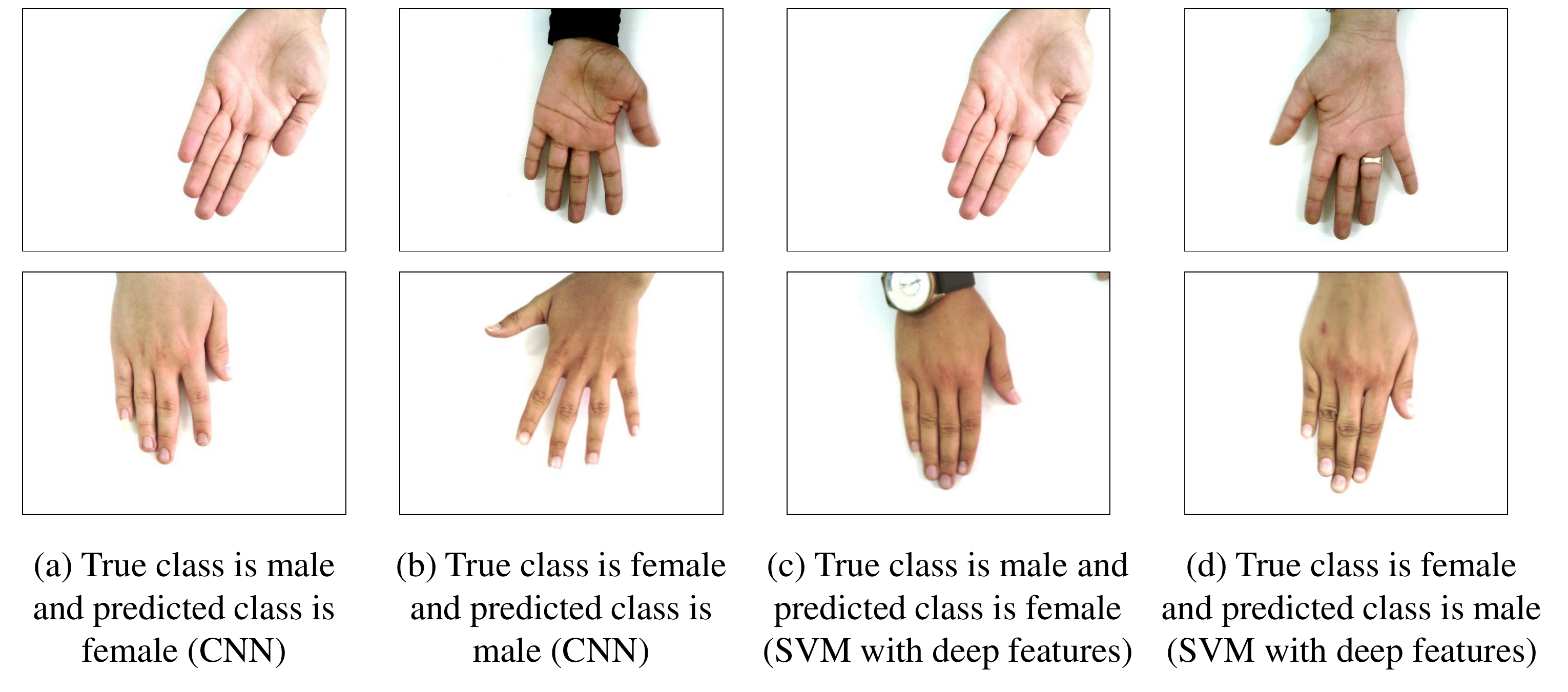}
   \caption{Failure cases of gender classification. Results obtained by the two-stream CNN, where the CNN misclassified the input hand image, are shown in (a) and (b). (c) and (d) show misclassified hand images by the SVM classifier with the deep features.}
    \label{fig:simialrity}
\end{figure}

\begin{table*}[]
\centering
\caption{Biometric identification results using our dataset. Each number ($N$-$S$) represents the average  identification accuracy obtained using $S$ side of the hand images for $N$ subjects.}
\label{Table3}

\begin{tabular}{|l|c|c|c|c|c|c|}
\hline
Method &  (80-P) &  (100-P) &  (120-P) &  (80-D)  &  (100-D) &  (120-D) \\ \hline

CNN-features/SVM & \textbf{0.948} & 0.929 & 0.933 & \textbf{0.964}  & \textbf{0.960} &\textbf{ 0.962}  \\ \hline

CNN-features+LBP/SVM & \textbf{0.960} & \textbf{0.953} & \textbf{0.956} & \textbf{0.962}  & \textbf{0.967} & \textbf{0.97}  \\ \hline
\end{tabular}

\end{table*}

\begin{table}[]
\centering
\caption{Biometric identification results using IITD dataset \cite{kumarincorporating}. Each number represents the identification accuracy using the corresponding number of subjects ($S$).}
\label{Table4}
\begin{tabular}{|l|c|c|c|c|}
\hline
Method & Features & $S=230$ & $S=137$ &  $S=100$ \\ \hline

Kumar and Shekhar \cite{kumar2010}& Gabor & \textbf{0.95 }& - & - \\ \hline
Kumar and Shekhar \cite{kumar2010} & Radon transform &  0.928 & - & - \\ \hline
Charfi \textit{et al.} \cite{charfi2014novel} & SIFT & 0.94 & - & - \\ \hline
Bera \textit{et al.} \cite{Bera2017} & finger contour profile &  \textbf{0.952} & \textbf{0.978} & -\\ \hline
Charfi \textit{et al.} \cite{new2017} & standard SIFT & 0.803 & - & 0.852 \\ \hline
Charfi \textit{et al.} \cite{new2017} & SIFT+SR & - & - & \textbf{ 0.962} \\ \hline

Ours & CNN-features & 0.9 & 0.912 & 0.891  \\ \hline

Ours & CNN-features+LBP& \textbf{0.948} &\textbf{0.964} & \textbf{0.94}  \\

 \hline
\end{tabular}

\end{table}

\begin{figure}
\centering
 \includegraphics[width=\linewidth]{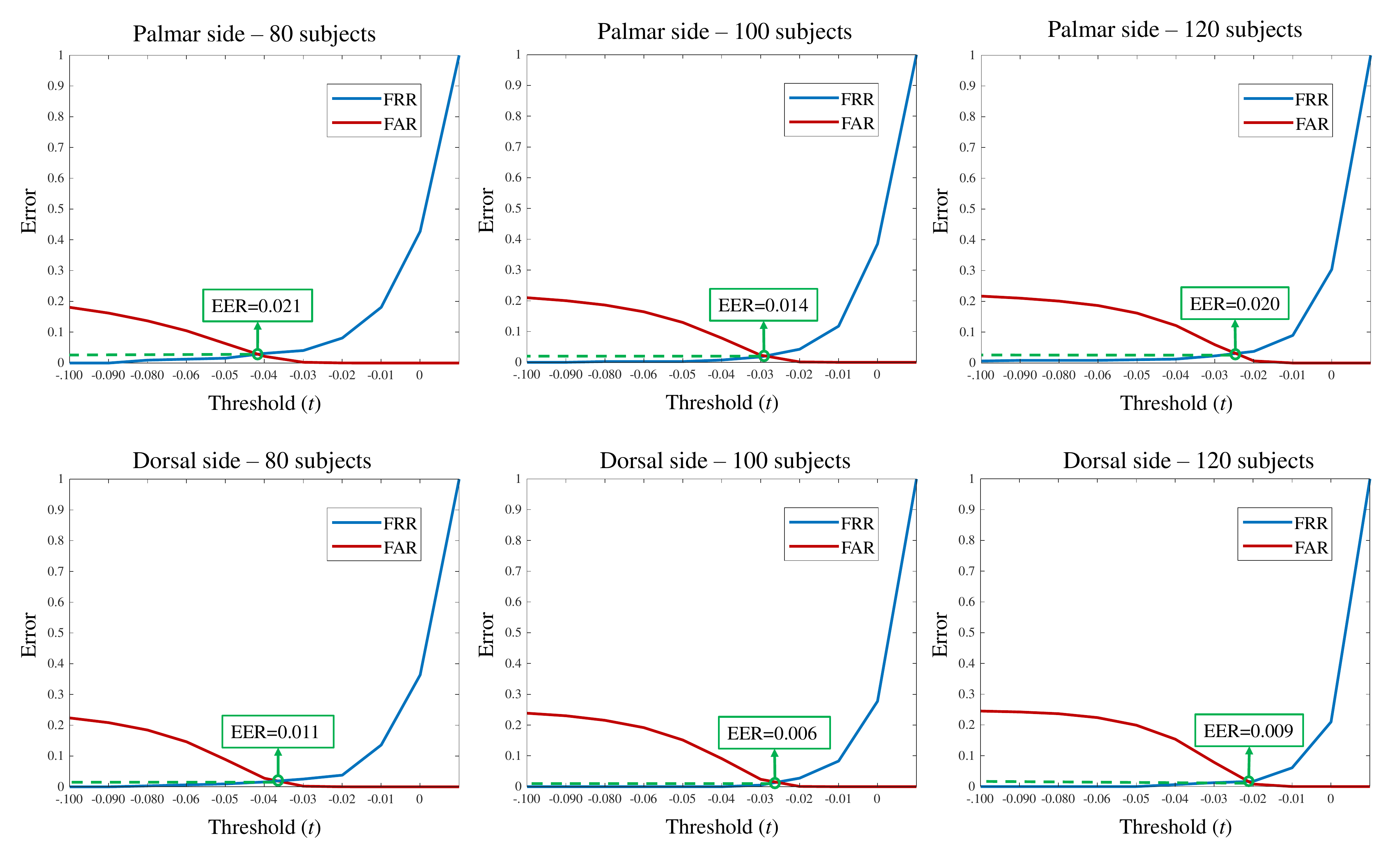}
   \caption{False acceptance rate (FAR), false rejection rate (FRR), and equal error rate (EER) for different population sizes of the 11K Hands dataset using the SVM classifier on the deep and LBP features.}
    \label{fig:id_analysis}
\end{figure}

\subsection{Error analysis}
For more understanding of the accuracy of our results on gender classification, we reported in Table \ref{Table6} the average confusion matrix of our trained networks on the 11K Hands dataset. The diagonal values are the classification accuracy for each gender; while the off-diagonal values are the misclassification rates. As shown, the ratio of misclassified hand images, for both genders, is reduced after using the SVM classifier with the deep features extracted from the trained two-stream CNN. 
\\
Fig. \ref{fig:simialrity} shows examples of the failure cases, where the trained classifier (either CNN or SVM with deep features) predicts wrong gender class. 
\\
On the other hand, there are three different metrics used to analysis the error of biometric identification algorithms. The first two metrics are the false acceptance rate (FAR) and false rejection rate (FRR). Both metrics indicates the likelihood of accepting an unauthorized user (i.e., FAR) or rejecting an authorized user (i.e., FRR). The last metric is the equal error rate (EER) which refers to the ratio of false rejections is equal to the ratio of false acceptances. Clearly, the EER is in an inverse proportion to the robustness of the biometric system. 
\\
In our case, we use one-against-all SVM classifier to classify the given hand image's features (deep and LBP features). In order to measure the FAR, FRR, and EER, we computed the SVM score for the given hand image's feature of being belongs to every class we have (i.e., subjects). Then, instead of assigning the given hand image to the subject whose score is the maximum (as we reported in Table \ref{Table3}), we assign the hand image to the subject if the score exceeds a certain threshold $t$. If this level is not reached for all subjects, the hand image is rejected. In other words, we validate each testing hand image against all subjects, including the true subject. The subjects and their hand images are picked randomly for each experiment.
\\
The FRR, FAR, and EER are reported in Fig. \ref{fig:id_analysis}. The equilibrium point is the EER. The range of the value of $t$ is from $\log(0.9)$ to $\log(1)$ with $0.01$ step -- the score of the SVM classifier is the $\log$ of the posterior probability. 

\begin{table}[]
\centering
\caption{Confusion matrices of the proposed method. The diagonal values are the true classification rate; while the off-diagonal values are the misclassification rates. The first two confusion matrices represents the results obtained using the palmar hand images using the CNN and SVM with the deep features, respectively. The second group of confusion matrices show results obtained using the dorsal side of hand images. For each confusion matrix, the row represents the predicted class and the column represents the true class.}
\label{Table6}
\begin{tabular}{l|l|l|l|l|l|l|l|l|}
\cline{2-9}
\multicolumn{1}{c|}{} & \multicolumn{2}{c|}{CNN} & \multicolumn{2}{c|}{\begin{tabular}[c]{@{}c@{}}SVM \\ (deep features)\end{tabular}} & \multicolumn{2}{c|}{CNN} & \multicolumn{2}{c|}{\begin{tabular}[c]{@{}c@{}}SVM \\ (deep features)\end{tabular}} \\ \cline{2-9} 
 & \multicolumn{4}{c|}{Palm side} & \multicolumn{4}{c|}{Dorsal side} \\  
 \cline{2-9}
 & Male & Female & Male &  Female & Male & Female & Male & Female \\ \hline
\multicolumn{1}{|l|}{Male} & \cellcolor[HTML]{EFEFEF}0.912 & 0.088 & \cellcolor[HTML]{EFEFEF}0.956 & 0.044 & \cellcolor[HTML]{EFEFEF}0.919 & 0.081 & \cellcolor[HTML]{EFEFEF}0.98 & 0.02 \\ \hline
\multicolumn{1}{|l|}{Female} & 0.165 & \cellcolor[HTML]{EFEFEF} 0.835 & 0.073 & \cellcolor[HTML]{EFEFEF}0.927 & 0.102 & \cellcolor[HTML]{EFEFEF}0.898 & 0.034 & \cellcolor[HTML]{EFEFEF}0.966 \\ \hline
\end{tabular}
\end{table}

\begin{figure}
\centering
 \includegraphics[width=\linewidth]{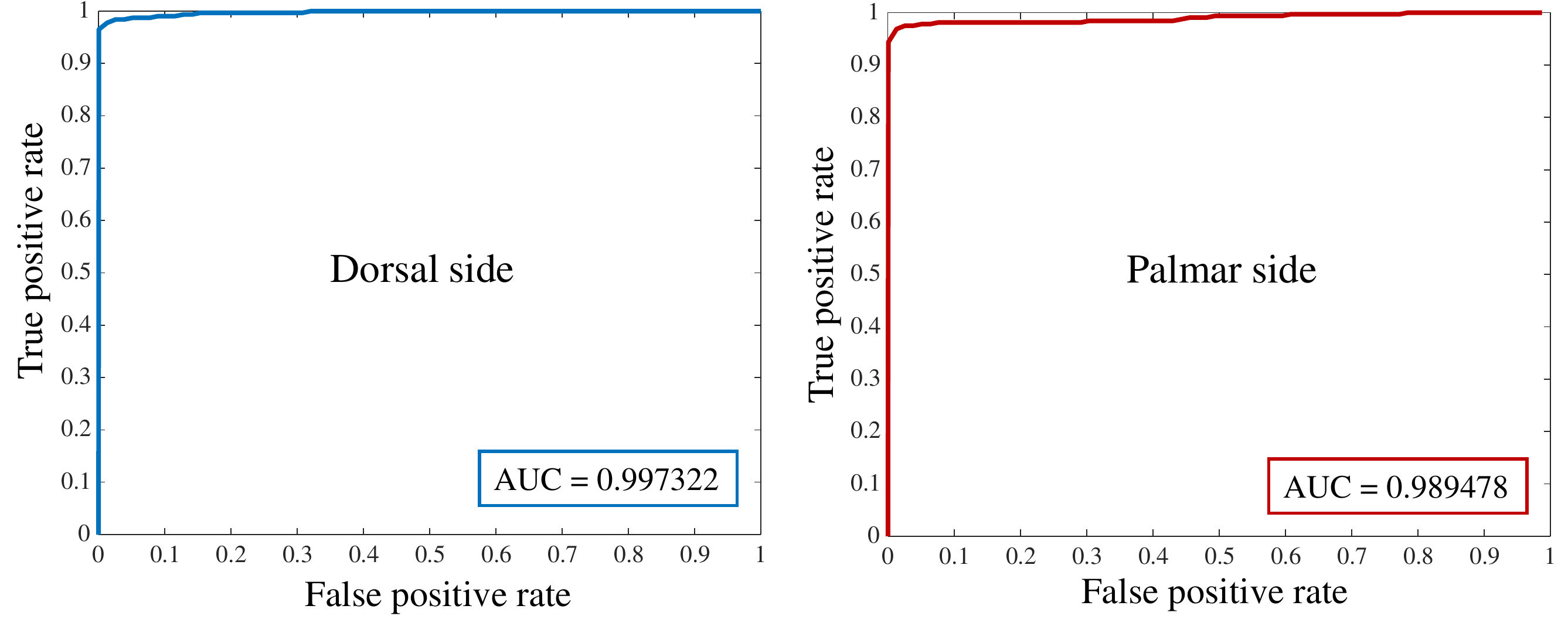}
   \caption{The receiver operating characteristic (ROC) curve for biometric identification of 80 subjects for both dorsal and palmar hand images. The term AUC denotes the area under the curve.}
    \label{fig:ROC}
\end{figure} 

\begin{figure}
\centering
 \includegraphics[width=\linewidth]{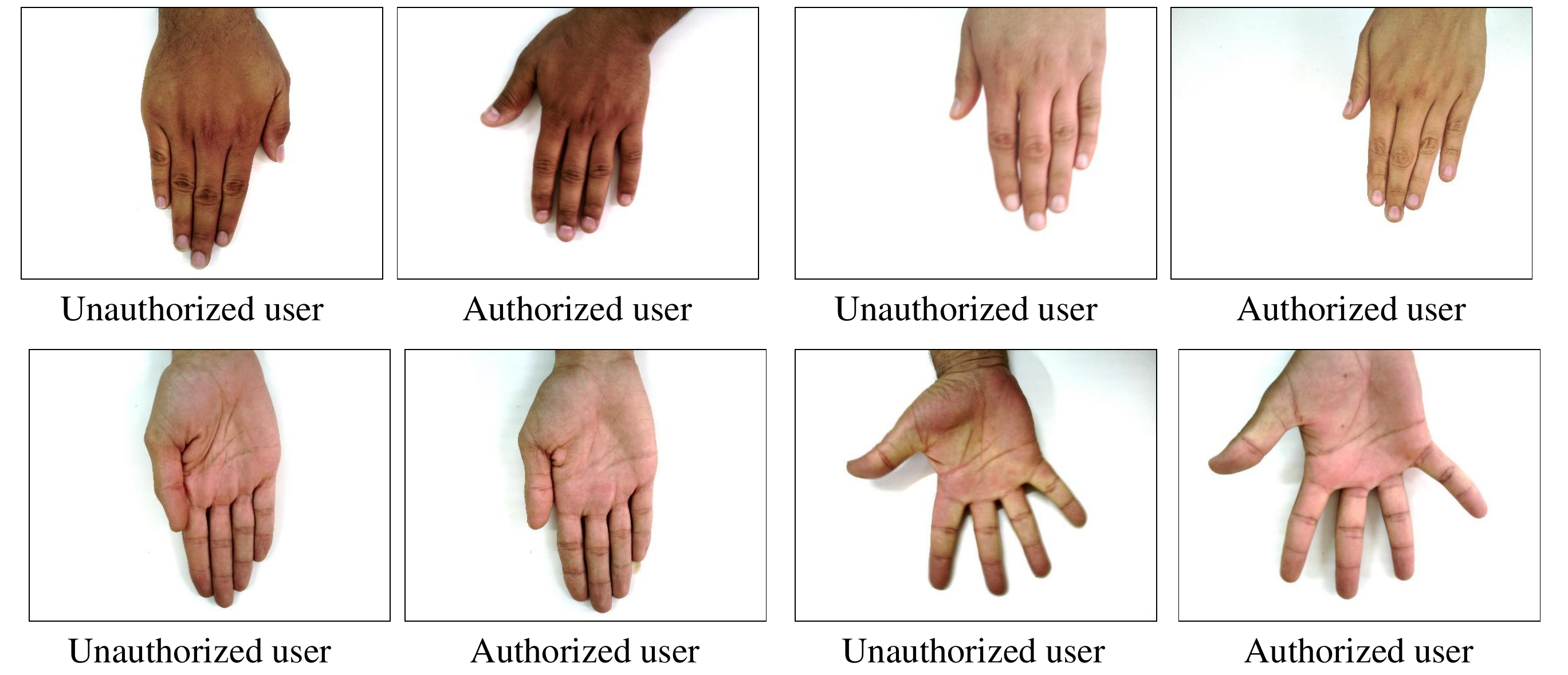}
   \caption{Failure cases of biometric identification. Each pair represents the unauthorized subject (left) and the authorized hand image (right).}
    \label{fig:failureCasesID}
\end{figure}

\noindent
The receiver operating characteristic (ROC) curve, shown in Fig. \ref{fig:ROC}, shows the true positive rate (TPR), also known as the sensitivity, against the false positive rate (FPR) at various threshold values for both dorsal and palmar hand images for 80 subjects -- the subjects and images picked randomly each time. Lastly, Fig. \ref{fig:failureCasesID} shows examples of unauthorized hand images that were incorrectly accepted by the classifier. As shown, there is a high similarity between each pair of images, namely, the reference hand image of the authorized subject and the unauthorized hand image (i.e., impostor user).

\section*{Reproducibility}
The proposed dataset, trained CNN models, and Matlab source code are publicly available at (\href{https://goo.gl/rQJndd}{https://goo.gl/rQJndd}) to encourage reproducibility of results.

\section{Conclusion}
\label{conclusion}
In this work, we have presented a new dataset of hand images for biometric identification and gender recognition. We have provided a set of useful metadata of the proposed dataset for each hand image. We also tested a set of state-of-art techniques on the dataset for gender classification. In addition, we presented a two-stream CNN as a base model for the proposed dataset for the gender classification problem. This CNN is used as a feature extractor that feeds a set of support vector machine classifiers for the biometric identification problem. 
\\
We have shown that dorsal hand images possesses distinctive features that could help in gender recondition and biometric identification problems. Where, the proposed method, bag of visual words and Fisher vector frameworks with different feature descriptors, AlexNet, VGG, and GoogleNet CNNs achieve better gender recognition accuracies using dorsal hand images comparing with the results obtained using the palmar side of hand images. In biometric identification, our method achieves better identification accuracy with dorsal side of hand images. 
\\
Extensive experiments showed that our method consistently attains better or on-par accuracy with state-of-the-art methods for both gender recognition and biometric identification.
\\
We believe that the proposed dataset can serve as a step towards the construction of more accurate gender classification and biometric identification systems that rely on hand images.


\end{document}